\documentclass[10pt,twocolumn,letterpaper]{article}

\usepackage[pagenumbers]{cvpr} %

\definecolor{cvprblue}{rgb}{0.21,0.49,0.74}
\usepackage[pagebackref,breaklinks,colorlinks,allcolors=cvprblue]{hyperref}

\usepackage{graphicx}
\usepackage{amsmath}
\usepackage{amssymb}
\usepackage{booktabs}
\usepackage{algorithm}
\usepackage{dsfont}
\usepackage{algorithm}
\usepackage{algpseudocode}
\usepackage{algorithmicx}
\usepackage{mathtools}
\usepackage{multirow}
\usepackage{graphics}
\usepackage{wrapfig}
\usepackage{bbm}
\usepackage{xspace}
\usepackage{xcolor}
\usepackage{tikz}

\usepackage{xr}
\makeatletter
\newcommand*{\addFileDependency}[1]{%
  \typeout{(#1)}
  \@addtofilelist{#1}
  \IfFileExists{#1}{}{\typeout{No file #1.}}
}
\makeatother

\def\paperID{3} %
\def\confName{CoRL}
\def\confYear{2025}

\title{FreeAction: Training-Free Techniques for Enhanced Fidelity of Trajectory-to-Video Generation}

\author{Seungwook Kim$^1$ \hspace{12mm} Seunghyeon Lee$^2$ \hspace{12mm}  Minsu Cho$^{1,3}$ \vspace{0.1cm}\\
POSTECH$^1$ \hspace{12mm} Ewha Womans University$^2$ \hspace{12mm} RLWRLD$^3$
}

\begin{document}

\twocolumn[{
\renewcommand\twocolumn[1][]{#1}%
\maketitle
\vspace{-8.0mm}
\includegraphics[width=1.0\linewidth]{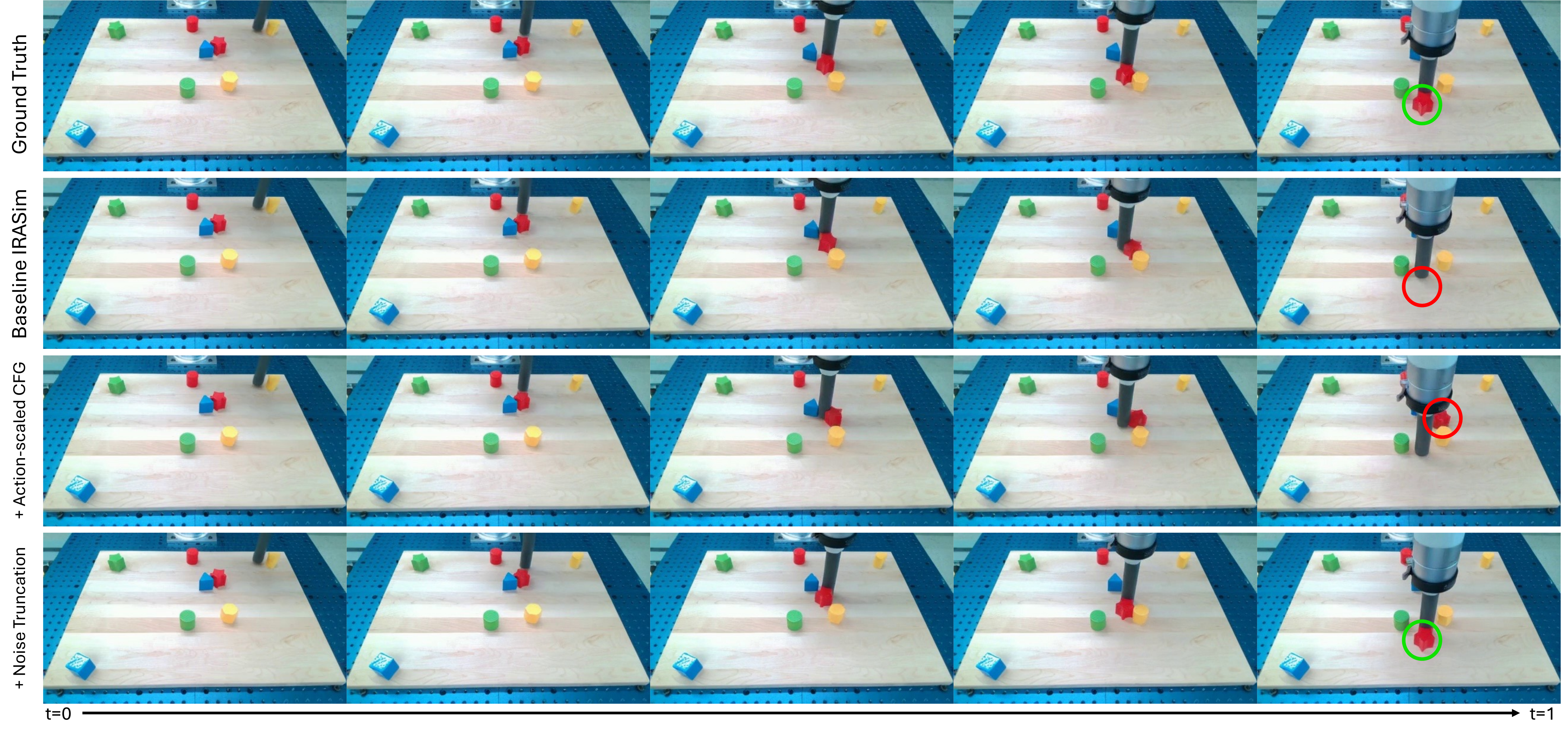}
\vspace{-7.5mm}
\captionof{figure}{\textbf{Results on the LanguageTable dataset}. 
      It can be seen that the red star block disappears in the baseline. 
      By integrating both our action-scaled CFG and noise truncation, it can be seen that the final output is visually identical to the ground-truth video.
}
\vspace{+4.0mm}
\label{fig:teaser}
}]

\begin{abstract}

Generating realistic robot videos from explicit action trajectories is a critical step toward building effective world models and robotics foundation models.
We introduce two training-free, inference-time techniques that fully exploit explicit action parameters in diffusion-based robot video generation.
Instead of treating action vectors as passive conditioning signals, our methods actively incorporate them to guide both the classifier-free guidance process and the initialization of Gaussian latents.
First, action-scaled classifier-free guidance dynamically modulates guidance strength in proportion to action magnitude, enhancing controllability over motion intensity.
Second, action-scaled noise truncation adjusts the distribution of initially sampled noise to better align with the desired motion dynamics.
Experiments on real robot manipulation datasets demonstrate that these techniques significantly improve action coherence and visual quality across diverse robot environments.

\end{abstract}

\section{Introduction}
\label{sec:introduction}

Generating robot videos conditioned on explicit action trajectories is a crucial component for building effective world models and robotics foundation models~\citep{du2023unipi,yang2023unisim,wu2024ivideogpt,irasim2024}.
By simulating robot trajectories across diverse environments and robots, such systems not only facilitate rapid prototyping and validation of control policies, but can also be used to generate rich synthetic data for downstream tasks such as policy learning and zero‑shot transfer~\citep{jiang2025enerverse, yang2025orv, fu2025learning, feng2025generalist, hu2024video}.

\begin{figure*}[ht]
    \begin{center}
        \includegraphics[width=1.0\linewidth]{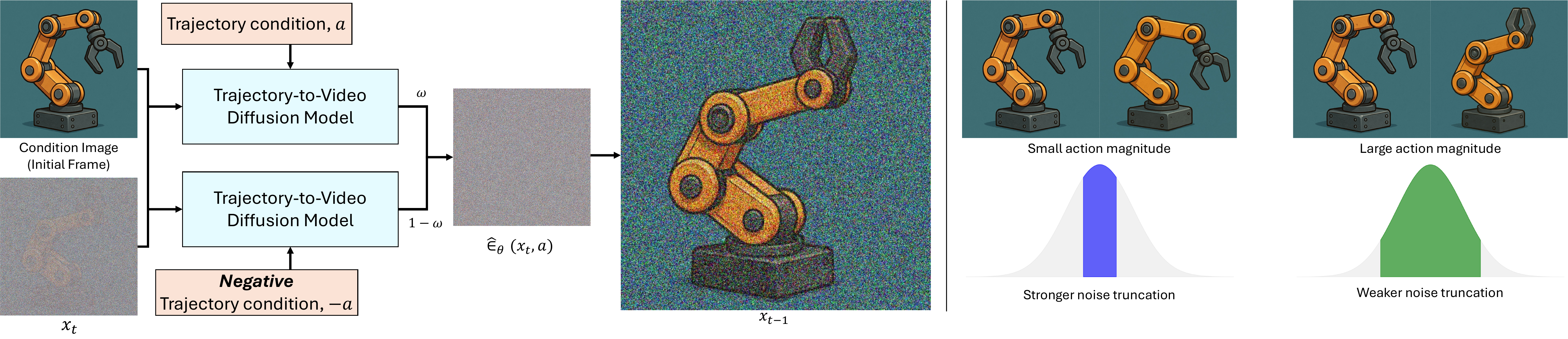}
    \end{center}
    \vspace{-5.0mm}
      \caption{\textbf{Illustration of our training-free inference-time techniques}. 
      (Left) Our action-scaled classifier-free guidance to steer the generation away from the negative trajectory, and towards our desired trajectory.
      We set $\omega = \lambda \|a\|_2\,\mathbf{1}_{\{t > T/2\}}$ so that the guidance strength varies in proportion to the action magnitude, and the guidance is applied only in the earlier steps. 
      (Right) Our action-scaled noise truncation scheme, where we manipulate the strength of truncation based on the magnitude of action.
}
\vspace{-3mm}
\label{fig:main_overview}
\end{figure*}

A recent line of work builds on diffusion-based generative models~\citep{peebles2023scalable, hong2022cogvideo, kim2024multi,kim2024enhancing} to produce robot videos with high visual fidelity~\citep{irasim2024, guo2024prediction}.
However, such existing trajectory-conditioned models~\citep{irasim2024} treat action trajectories as passive conditioning signals \textit{e.g.,} injecting the action trajectories as conditions into the diffusion process via Ada-LN~\citep{huang2017arbitrary} in a Diffusion Transformer (DiT)\citep{peebles2023scalable} architecture.
In this paper, we posit that action trajectories are currently underutilized and can be leveraged more effectively to enhance trajectory coherence in generated robot videos.

To address this gap, we present two training-free, inference-time techniques that actively exploit action trajectory information to steer the diffusion process toward improved action coherence in generated videos.
The first modulates classifier-free guidance strength according to action magnitude, while the second manipulates initial noise sampling to enhance the fidelity and determinism of the trajectory-to-video generation process, reflecting the intended motion dynamics.

Our contributions are as follows:
\begin{itemize}
    \item We propose \emph{action-scaled classifier-free guidance}, which dynamically adjusts the guidance weight in proportion to the action magnitude, improving trajectory coherence.
    \item We introduce \emph{action-scaled noise truncation} that modifies the initially sampled Gaussian noise to better align generated motion with the specified action parameters.
    \item We demonstrate on three real-robot manipulation datasets that our methods yield substantial improvements in both action fidelity and visual quality.
\end{itemize}

\section{Action-scaled CFG and Noise Truncation}
\label{sec:method}

\smallbreak
\noindent
\textbf{Preliminary: IRASim}.
IRASim~\citep{irasim2024} is a DiT~\citep{dit2022}-based action trajectory-to-video generative diffusion model, which functions as an interactive real-robot simulator.
In essence, given an initial frame $I^{1}$ and an action trajectory ${a}^{1:N-1}$ as conditions, IRASim generates $N-1$ future frames $I^{1:N}$ which reflect the results of the action trajectory on the initial frame.
IRASim injects action trajectories as conditions via the AdaLN layers~\citep{huang2017arbitrary}.
In the following, we introduce how we analyze and manipulate the conditioning and noise initialization mechanisms in the inference pipeline of IRASim to improve the visual quality and action coherency of generated videos. 

\begin{table*}[ht]
  \centering
  \caption{\textbf{Quantitative results on RT-1, Bridge, and Language-Table datasets.} 
    Across both short- and long-trajectory evaluation settings, using our test-time techniques improves results compared to the baseline IRASim~\citep{irasim2024}.
    }
    \label{tab:main_quantitative}
    \vspace{-3.0mm}
  \begin{tabular}{l l cc c}
    \toprule
    Dataset          & Method                  &  PSNR$\uparrow$   & SSIM$\uparrow$    & Latent L2$\downarrow$  \\
    \midrule
    \multicolumn{5}{l}{\textit{Short-trajectory}} \\
    \midrule
    \multirow{3}{*}{RT‑1~\citep{brohan2022rt}}
                    & {IRASim‑Frame‑Ada}       & 26.024  & 0.833    & 0.2100 \\                  
                    & + Action-scaled CFG       & \underline{26.198}  & \underline{0.837}    & \underline{0.2068}  \\
                    & + Action-scaled Noise Truncation       & \textbf{26.435}  & \textbf{0.840}    & \textbf{0.1629} \\
    \midrule
    \multirow{3}{*}{Bridge~\citep{walke2023bridgedata}}
                    & {IRASim‑Frame‑Ada}       & 25.340  & 0.834    & 0.1939  \\                  
                    & + Action-scaled CFG       & \underline{25.398}  & \underline{0.835}    & \underline{0.1938} \\
                    & + Action-scaled Noise Truncation       & \textbf{25.770}  & \textbf{0.843}    & \textbf{0.1662} \\
    \midrule
    \multirow{3}{*}{Language‑Table~\citep{lynch2023interactive}}
                    & {IRASim‑Frame‑Ada}       & 28.794  & 0.888    & 0.1663  \\    
                    & + Action-scaled CFG       & \underline{29.021}  & \underline{0.890}    & \underline{0.1653}  \\
                    & + Action-scaled Noise Truncation       & \textbf{29.514}  & \textbf{0.902}    & \textbf{0.1326}  \\

    \midrule
    \multicolumn{5}{l}{\textit{Long-trajectory}} \\
    \midrule
    \multirow{3}{*}{RT‑1~\citep{brohan2022rt}}
                    & {IRASim‑Frame‑Ada}       &  21.729 &  0.760   & 0.2408 \\                  
                    & + Action-scaled CFG       &  \underline{21.984} & \underline{0.763}   & \underline{0.2355} \\
                    & + Action-scaled Noise Truncation       & \textbf{22.299} & \textbf{0.776} & \textbf{0.1891} \\
    \midrule
    \multirow{3}{*}{Bridge~\citep{walke2023bridgedata}}
                    & {IRASim‑Frame‑Ada}       & 21.536 & 0.769 & 0.2306  \\                  
                    & + Action-scaled CFG       & \underline{21.601}  & \underline{0.771}   & \underline{0.2302} \\
                    & + Action-scaled Noise Truncation       & \textbf{22.035} & \textbf{0.785} & \textbf{0.1961} \\
    \midrule
    \multirow{3}{*}{Language‑Table~\citep{lynch2023interactive}}
                    & {IRASim‑Frame‑Ada}       & 24.861 & 0.852 & 0.1730  \\    
                    & + Action-scaled CFG       & \underline{24.920} & \underline{0.853} & \underline{0.1719} \\
                    & + Action-scaled Noise Truncation & \textbf{25.120} & \textbf{0.867} & \textbf{0.1394} \\
    \bottomrule
  \end{tabular}
  \vspace{-3mm}
\end{table*}

\begin{figure*}[ht]
    \begin{center}
        \includegraphics[width=0.9\linewidth]{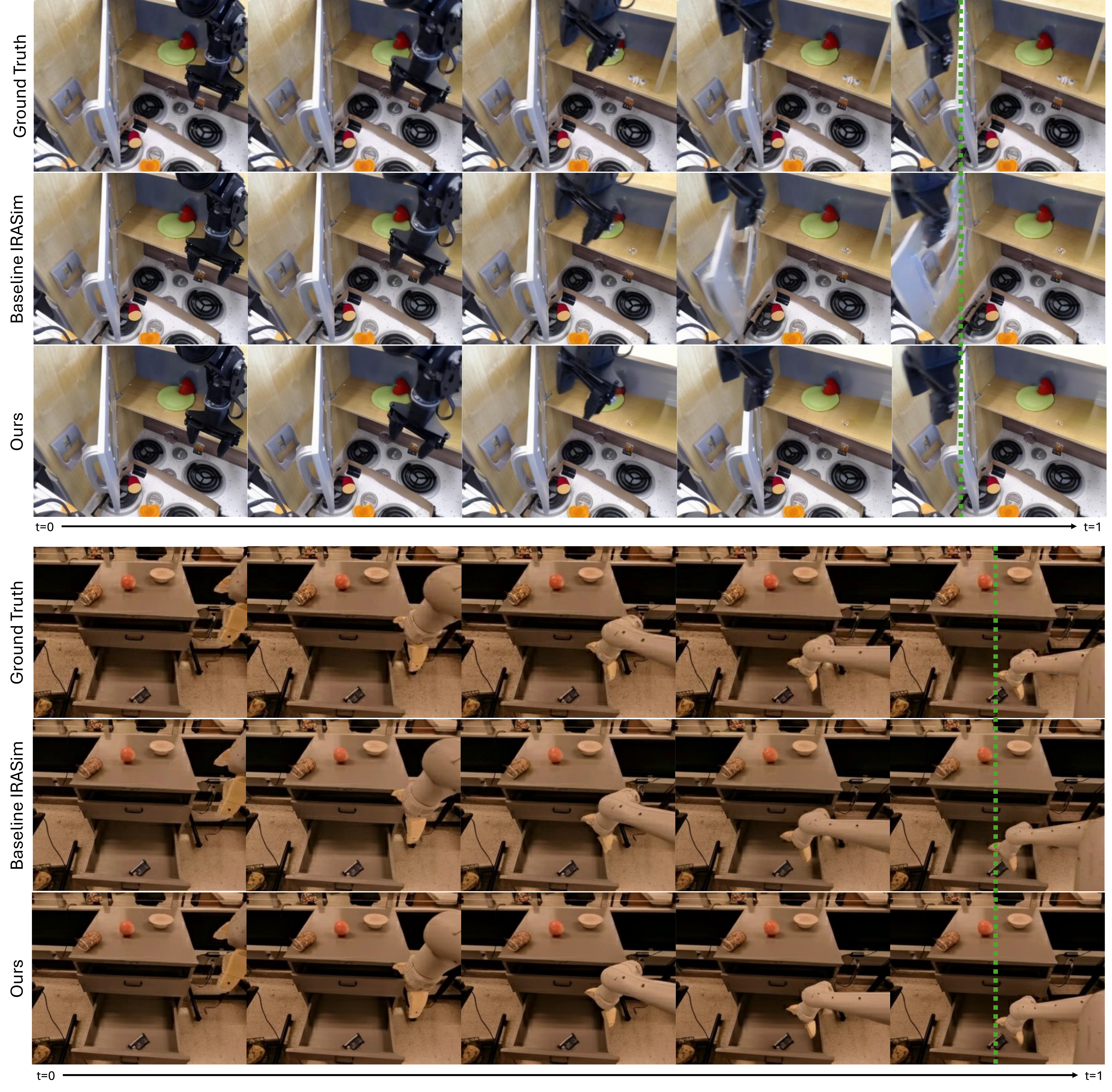}
    \end{center}
    \vspace{-5mm}
      \caption{\textbf{Qualitative results on the Bridge (top) and RT-1 (bottom) datasets}. 
      Applying our techniques improves coherency to the action trajectory.
      Note the green line, where it can be seen that ours generates results which better resemble the ground truth video.
}
\vspace{-3.5mm}
\label{fig:main_qual}
\end{figure*}

\subsection{Action-scaled Classifier-free Guidance}

\smallbreak
\noindent
Classifier-Free Guidance (CFG)~\citep{ho2022classifier} is a standard method for improving the fidelity of diffusion-generated outputs.
In text-to-image models, CFG is implemented by randomly dropping text prompts during training, enabling the diffusion model to act as both a conditional and an unconditional denoiser.
Given a text prompt $c$, and an unconditional token $\varnothing$, the standard CFG update at step $t$ is as follows: 
\begin{equation}
\hat{\epsilon}\theta(x_t, c)
=
\epsilon\theta(x_t, \varnothing)
+
\omega\left[\epsilon_\theta(x_t, c) - \epsilon_\theta(x_t, \varnothing)\right],
\label{eq:cfg-classic}
\end{equation}
where $\omega=0$ recovers unconditional prediction, $\omega=1$ recovers purely conditional prediction, and $\omega > 1$ amplifies conditioning.
However, this "prompt-dropping" strategy substantially increases the training complexity~\citep{hong2023improving, ahn2024self}, and directly extending prompt-dropping to action-to-video models introduces semantic inconsistencies, since a zero action should represent no motion rather than a neutral prior.

To overcome these challenges, we propose \textbf{action‑scaled classifier‑free guidance}.
Instead of learning an unconditional denoiser, we synthesize a \textit{negative} action condition by {negating} the input action, \textit{i.e.,} $-a$.
Concretely, if $\epsilon_{\theta}(x_t, a)$ denotes the predicted noise given action $a$, and $\epsilon_{\theta}(x_t, -a)$ the prediction given the negated action, our guided update at timestep $t$ takes the form:
\begin{equation}
    \hat{\epsilon}_\theta(x_t, a) \;=\; \epsilon_{\theta}(x_t, a)
    \;+\; \omega\,\bigl[\epsilon_{\theta}(x_t, a) \;-\; \epsilon_{\theta}(x_t, -a)\bigr].
\end{equation}
Intuitively, this encourages samples to move closer to the trajectory specified by $a$ and farther from its negation.

However, two practical issues remain in the above formulation: (1) small-magnitude actions should result in proportionally weaker repulsion from the negative condition, and (2) the high-frequency texture details, which are largely independent of motion, should be preserved.
We address both by scaling the guidance weights as $\omega = \lambda \|a\|_2\,\mathbf{1}_{\{t > T/2\}}$, where $|a|_2$ is the $\ell_2$-norm of the action vector, $t$ is the current step, and $T$ is the total number of sampling steps. 
$\lambda$ is a hyperparameter\footnote{We set $\lambda=1$ in our experiments.}. 
This formulation increases guidance strength for larger actions, minimizes perturbations for small ones, and applies the effect only in early steps (low-frequency generation) to maintain texture fidelity in later refinements.
We illustrate our action-scaled classifier-free guidance in the left of Fig.~\ref{fig:main_overview}.

\subsection{Action-scaled Noise Truncation}

In conventional text-to-video generation, models are expected to balance fidelity and diversity.
In contrast, in the current task of trajectory-to-video, the goal is a single faithful realization of the given specific action trajectory $a$.
We therefore propose to regulate the diversity of generated videos by manipulating the initial Gaussian latent $\mathbf{z}_0 \sim \mathcal{N}(\mathbf{0},\mathbf{I})$, via action-scaled noise truncation, in the spirit of BigGAN’s truncation trick~\cite{brock2018large}.
This truncation constrains sampling to high-density regions, yielding higher fidelity while sacrificing diversity - a trade-off that is desirable for trajectory-to-video.

We implement min–max truncation with a sigmoid mapping from action magnitude to an element-wise truncation limit, centered at the dataset mean:
\begin{equation}
\begin{aligned}
\tau(a) &= \tau_{\min} + (\tau_{\max}-\tau_{\min})\,
  \sigma\!\big(\lVert a\rVert_2 - \mu_{\text{act}}\big),
\end{aligned}
\label{eqn:action_scaled_noise_trunction}
\end{equation}
where $\sigma()$ is the sigmoid function, $\mu_{\text{act}}$ is the dataset mean of $\lVert a\rVert_2$\footnote{We pre-calculate the mean each dataset from their train set.} and $0<\tau_{\min}\le \tau_{\max}$\footnote{We set $\tau_{\min}=0.5$ and $\tau_{\max}=1.5$ in our experiments.}. 
We then draw the initial latent by sampling each entry independently from a zero-mean truncated normal with limit $\tau(a)$: 
$$ 
z_{0,i} \sim \mathcal{N}(0,1)\ \text{conditioned on}\ |z_{0,i}|\le \tau(a)\quad\forall i
$$
Small actions ($\lVert a\rVert_2 \ll \mu_{\text{act}}$) induce strong truncation $(\tau \approx \tau_{\min})$ to aim for nearly deterministic appearance; large actions relax truncation $(\tau \approx \tau_{\max})$ to retain the variation needed to express substantial motion and appearance change from the previous frame.
This plug-and-play modification requires no additional training.
We illustrate our action-scaled noise truncation in the right of Fig.~\ref{fig:main_overview}.

\section{Experimental Results}
\label{sec:experiments}

\smallbreak
\noindent
\textbf{Evaluation setup.} Following IRASim~\citep{irasim2024}, we evaluate our method on the real-robot manipulation datasets of (1) RT-1~\citep{brohan2022rt} with 7 DoF action space, (2) Bridge~\citep{walke2023bridgedata} with 7 DoF action space, and (3) LanguageTable~\citep{lynch2023interactive} with 2 DoF action space.
In short-trajectory evaluation (single generation pass), we use a single frame as reference and 15 actions to predict the next 15 frames.
For the long-trajectory evaluation, we use a single frame as reference, and generate the video autoregressively where the final frame from the previous pass serves as the reference frame for the next pass.
Noting that Latent L2 loss and PSNR best align with human preferences~\citep{irasim2024}, we report Latent L2, PSNR and additionally SSIM to evaluate the performance of our method.

\smallbreak
\noindent
\textbf{Results and analyses.}
We show the results of our two training-free techniques in Table~\ref{tab:main_quantitative}, on both the short- and long-trajectory evaluation settings.

It can be seen that our proposed techniques, action-scaled CFG and action-scaled noise truncation shows consistent improvements over the baseline IRASim-Frame-Ada on all metrics of PSNR, SSIM and Latent L2, \ie, effectively achieving higher perceptual quality and fidelity on both settings of short- and long-trajectory.
We also provide qualitative results on the Language-Table dataset in Figure~\ref{fig:teaser}, and on the Bridge and RT-1 datasets in Figure~\ref{fig:main_qual}.
In both figures, we visually show that while the baseline method already shows photorealistic outputs, applying our techniques leads to improved action coherence, showing better similarity with the ground-truth video.
We provide additional experiments and analyses in the appendix.

\section{Discussion and Future Direction}
\label{sec:conclusion}

The proposed action-scaled classifier-free guidance and action-scaled noise truncation are training-free, inference-time controls that improve visual fidelity and action coherence in diffusion-based trajectory-to-video generation.
We demonstrated the efficacy of our techniques over real-robot RT-1, Bridge and Language-Table datasets.
We expect that these techniques will enable trajectory-conditioned generators to generate more coherent rollouts without having to modify or re-train the base model.
Looking ahead, we will investigate effective techniques in expanded conditioning regimes - \eg, models that accept \textit{natural language} instructions or additional modalities such as depth.

\clearpage
\noindent
\small
\textbf{Acknowledgement.} 
This work was supported by the IITP grants (RS-2022-II220959: Few-Shot Learning of Causal Inference in Vision and Language for Decision Making (50\%), RS-2025-25443730: Research and Evaluation Framework for Robotic Artificial General Intelligence with Broad Generalization Capabilities (45\%), RS-2019-II191906: AI Graduate School Program at POSTECH (5\%)) funded by the Korea government (MSIT).
{
    \small
    \bibliographystyle{ieeenat_fullname}
    \bibliography{main}
}

\clearpage
\appendix
\clearpage  %

\begin{table*}[h!]
  \centering
  \caption{\textbf{Ablation experiments on RT-1, Bridge, and Language-Table datasets.} 
    }
    \label{tab:supp_ablation}
  \begin{tabular}{l l cc c}
    \toprule
    Dataset          & Method                  &  PSNR$\uparrow$   & SSIM$\uparrow$    & Latent L2$\downarrow$  \\

    \midrule
    \multicolumn{5}{l}{\textit{vs. CFG using fixed guidance weight $\omega$}} \\
    \midrule
    \multirow{3}{*}{RT‑1~\citep{brohan2022rt}}
                    & $\omega=1.0$ (baseline)       &  26.024 & 0.833 & 0.2100 \\
                    & $\omega=3.0$        &  26.052 &  0.832  & 0.2112\\
                    & \textbf{Ours} ($\omega = \lambda \|a\|_2\,\mathbf{1}_{\{t > T/2\}}$)       &26.198 & 0.837 & 0.2068\\
    \midrule
    \multirow{3}{*}{Bridge~\citep{walke2023bridgedata}}
                & $\omega=1.0$ (baseline)       &  25.340 & 0.834 & 0.1939 \\
                    & $\omega=3.0$        & 25.175  & 0.830   & 0.1950 \\
                    & \textbf{Ours} ($\omega = \lambda \|a\|_2\,\mathbf{1}_{\{t > T/2\}}$)       &25.398 & 0.835 & 0.1938 \\
    \midrule
    \multirow{3}{*}{Language‑Table~\citep{lynch2023interactive}}
                    & $\omega=1.0$ (baseline)       &  28.794 & 0.888 & 0.1663 \\
                    & $\omega=3.0$        & 28.922  & 0.890   & 0.1654 \\
                    & \textbf{Ours} ($\omega = \lambda \|a\|_2\,\mathbf{1}_{\{t > T/2\}}$)       & 29.021 & 0.890 & 0.1653\\

    \midrule
    \multicolumn{5}{l}{\textit{vs. Fixed truncation level $\tau(a)$}} \\
    \midrule
    \multirow{3}{*}{RT‑1~\citep{brohan2022rt}}
                    & $\tau(a)=1.0$       & 26.370 & 0.840  & 0.1669\\
                    & $\tau(a)=1.5$        & 26.236 & 0.839  & 0.1892 \\
                    & \textbf{Ours} (Equation~\ref{eqn:action_scaled_noise_trunction})      &  26.435 & 0.840  & 0.1629 \\           
    \midrule
    \multirow{3}{*}{Bridge~\citep{walke2023bridgedata}}
                & $\tau(a)=1.0$       & 25.782 & 0.841 & 0.1690 \\
                    & $\tau(a)=1.5$        & 25.487 & 0.838  &  0.1766 \\
                    & \textbf{Ours} (Equation~\ref{eqn:action_scaled_noise_trunction})      &  25.770 & 0.843   & 0.1662 \\            
    \midrule
    \multirow{3}{*}{Language‑Table~\citep{lynch2023interactive}}
                    & $\tau(a)=1.0$       & 29.488 & 0.899 & 0.1369 \\
                    & $\tau(a)=1.5$        & 29.122  & 0.892    & 0.1459 \\
                    & \textbf{Ours} (Equation~\ref{eqn:action_scaled_noise_trunction})      &  29.514 & 0.902   & 0.1326 \\            
    \bottomrule
  \end{tabular}
\end{table*}
\begin{figure*}[htbp]
    \setlength{\abovecaptionskip}{0pt}
    \begin{center}
        \includegraphics[width=1.0\linewidth]{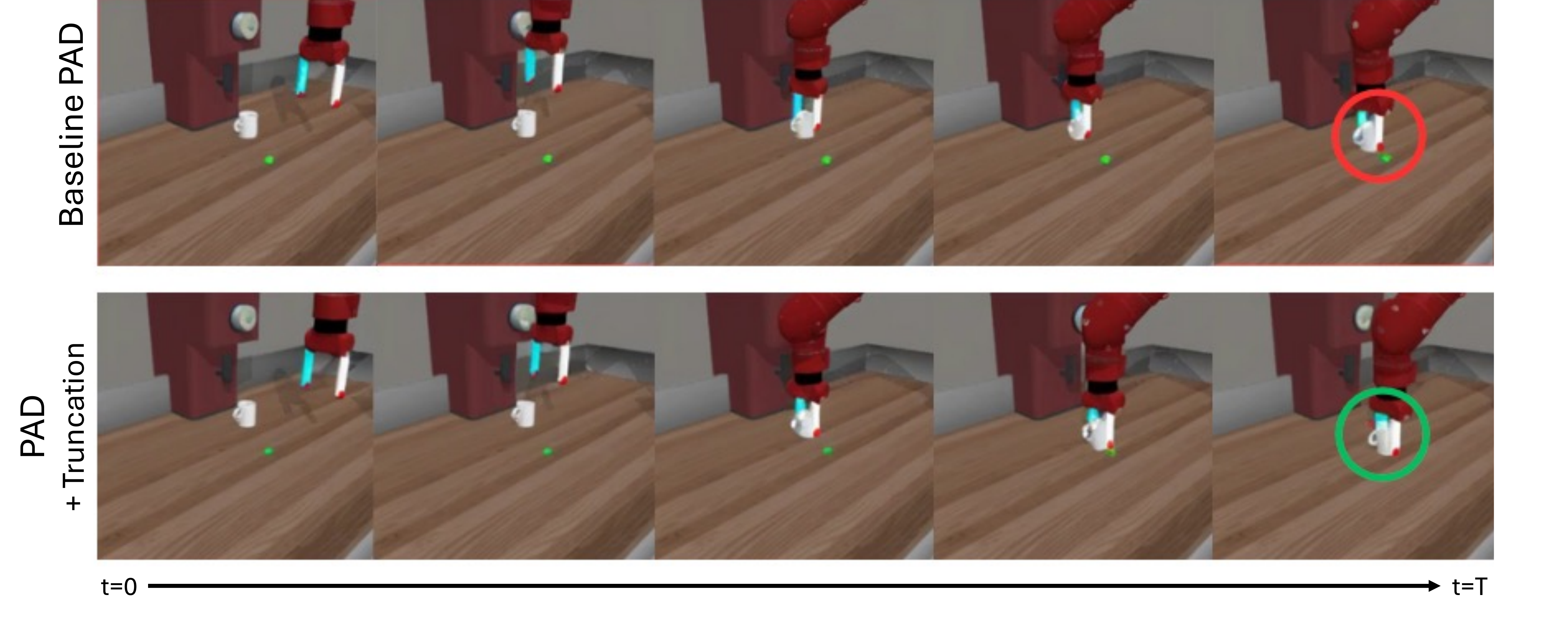}
    \end{center}
    \setlength{\belowcaptionskip}{0pt}
      \caption{\textbf{Qualitative example of PAD with truncation on the Metaworld benchmark.}. 
      The figure compares baseline PAD and PAD with truncation in the \texttt{coffee-pull-v2} task, where the cup is expected to be placed at the green dot to be counted as a success. 
      While the baseline often fails to place the coffee cup at the target green dot, PAD with truncation successfully reaches the target, demonstrating more consistent and accurate manipulation over time.
}
\label{fig:appendix_pad_qual}
\end{figure*}

\section{Comparative analysis}

We compare \textit{action-scaled} CFG and noise truncation against their \textit{fixed-weight} counterparts.
As summarized in Table~\ref{tab:supp_ablation}, the action-scaled variants achieve the best overall performance across datasets and metrics. 
Specifically, using a fixed CFG weight $(\omega=\text{const})$ can degrade results, likely due to miscalibrated steering, \ie, the guidance becomes too strong for small actions and too weak for large ones.
A fixed truncation threshold $(\tau=\text{const})$ consistently improves upon the IRASim baseline by reducing spurious variability in the initial latent; however, scaling the truncation $\tau(\mathbf{a})$ with the action norm yields the strongest gains by enforcing determinism for small motions while preserving necessary diversity for large motions.

\section{Application to Visual Policy Learning}
\begin{table*}[htbp]
\centering
\caption{\textbf{Success rate comparison under different input modality configurations on the MetaWorld benchmark}, with and without the handle-pull task (50-task benchmark).}
\label{tab:appendix_PAD}
\begin{tabular}{lcc}
\toprule
\textbf{Configuration} & \textbf{w/o handle-pull} ($\%$) & \textbf{w/ handle-pull} ($\%$) \\
\midrule
Baseline & 70.00 & 67.20 \\
Truncation at Image  & 70.42 & 67.60 \\
Truncation at Action, Depth, Image & 69.38 & 66.60 \\
Truncation at Depth, Image & \textbf{71.46} & \textbf{68.60} \\
\bottomrule
\end{tabular}
\label{tab:supp_pad}
\end{table*}

We test whether our insights transfer beyond trajectory-to-video generation. 
Specifically, our approach regulates sample diversity to improve fidelity; here, we evaluate the \emph{noise truncation} component on a visual policy learning task using Prediction with Action (PAD)~\citep{guo2024prediction} as the baseline. 
Because we could not find an open-source visual policy framework that conditions on action trajectories, and PAD conditions on natural language instructions instead, we disable CFG and evaluate a fixed-threshold truncation (\(\tau=1.0\)).

\smallbreak
\noindent
\textbf{Experimental setup (PAD).}
PAD~\citep{guo2024prediction} is a DiT-based diffusion model that jointly predicts future visual observations and robot actions. 
We evaluate on the MetaWorld benchmark~\citep{yu2020meta}, reporting results both \emph{including} and \emph{excluding} \texttt{handle-pull-v2}, which is known to be unusually difficult\footnote{Baseline PAD yields \(0\%\) success on \texttt{handle-pull-v2}.}. 
PAD supports three generative modalities—image, depth, and action—each initialized element-wise from \(\mathcal{N}(0,1)\). 
We apply fixed-threshold noise truncation to different modality subsets: (1) Image; (2) Action+Depth+Image; (3) Depth+Image.

\smallbreak
\noindent
\textbf{Results and analysis.}
Table~\ref{tab:supp_pad} shows that truncating the initial noise for the \emph{image} or \emph{depth} modalities consistently improves success rates over the PAD baseline. 
In contrast, truncating the \emph{action} modality harms performance: unlike image/depth, which are strongly correlated across frames, action vectors are not reliably autocorrelated\footnote{States (e.g., positions) are temporally correlated, but the \emph{actions} that transition between them need not be.}, so reducing action-sampling variability can over-constrain the next-action prediction. 
Overall, trading off diversity for fidelity in modalities with strong temporal correlation (image, depth) yields gains beyond trajectory-to-video generation, and—even without modifying the action channel—the increased determinism in image/depth positively influences action prediction, improving task success. 
Qualitative examples in Fig.~\ref{fig:appendix_pad_qual} illustrate more precise placements (e.g., moving the cup to the green target) when truncation is applied.

\end{document}